\pdfoutput=1
\documentclass[runningheads]{llncs}
\usepackage{graphicx}

\usepackage{tikz}
\usepackage{comment} 
\usepackage{amsmath,amssymb} 
\usepackage{color}

\usepackage[caption=false]{subfig}
\usepackage{booktabs}
\usepackage{multirow}
\usepackage{pifont}
\newcommand{\cmark}{\ding{51}}
\newcommand{\xmark}{\ding{55}}
\usepackage{algorithm}
\usepackage{algorithmic}
\usepackage[hidelinks]{hyperref}
\usepackage{url}
\usepackage{verbatim}
\usepackage[utf8]{inputenc}

\begin{document}
\pagestyle{headings}
\mainmatter
\def\ECCVSubNumber{701}  

\title{Webly Supervised Image Classification with \\ Self-Contained Confidence} 

\titlerunning{Webly Supervised Image Classification with Self-Contained Confidence}
%
\author{Jingkang Yang\inst{1,2}\thanks{Work done during an internship at SenseTime EIG Research.} \and
Litong Feng\inst{1} \and
Weirong Chen\inst{1,3\star} \and
Xiaopeng Yan\inst{1} \and \\
Huabin Zheng\inst{1} \and
Ping Luo\inst{4} \and
Wayne Zhang\inst{1}\orcidID{0000-0002-8415-1062}
}
\authorrunning{J. Yang et al.}
%

\institute{SenseTime Research\\
\email{\{yangjingkang,fenglitong,chenweirong,yanxiaopeng,\\zhenghuabin,wayne.zhang\}@sensetime.com}\\
\and Rice University, Houston, TX, USA\\
\and The Chinese University of Hong Kong, Hong Kong SAR, China\\
\and The University of Hong Kong, Hong Kong SAR, China}


\maketitle
	
\begin{abstract}
This paper focuses on webly supervised learning (WSL), where datasets are built by crawling samples from the Internet and directly using search queries as web labels. 
Although WSL benefits from fast and low-cost data collection, noises in web labels hinder better performance of the image classification model. 
To alleviate this problem, in recent works, self-label supervised loss $\mathcal{L}_s$ is utilized together with webly supervised loss $\mathcal{L}_w$. $\mathcal{L}_s$ relies on pseudo labels predicted by the model itself. 
Since the correctness of the web label or pseudo label is usually on a case-by-case basis for each web sample, it is desirable to adjust the balance between $\mathcal{L}_s$ and $\mathcal{L}_w$ on sample level. 
Inspired by the ability of Deep Neural Networks (DNNs) in confidence prediction, we introduce Self-Contained Confidence (SCC) by adapting model uncertainty for WSL setting, and use it to sample-wisely balance $\mathcal{L}_s$ and $\mathcal{L}_w$. 
Therefore, a simple yet effective WSL framework is proposed. A series of SCC-friendly regularization approaches are investigated, among which the proposed graph-enhanced mixup is the most effective method to provide high-quality confidence to enhance our framework. The proposed WSL framework has achieved the state-of-the-art results on two large-scale WSL datasets, WebVision-1000 and Food101-N. Code is available at \hyperlink{https://github.com/bigvideoresearch/SCC}{https://github.com/bigvideoresearch/SCC}.
\end{abstract}

\section{Introduction}

Large-scale human-labeled data plays a vital role in deep learning-based applications such as image classification~\cite{deng2009imagenet}, scene recognition~\cite{zhou2017places}, face recognition~\cite{sun2014deep}, etc. However, high-quality human annotations require significant cost in labor and time. Webly supervised learning (WSL), therefore, has attracted more attention recently as a cost-effective approach for developing learning systems from abundant web data. Generally, search queries fed into image crawlers are directly used as web labels for crawled images, which also introduce label noise due to semantic ambiguity and search engine bias. How to deal with these unreliable and noisy web labels becomes a key task in WSL.

\begin{figure}[t]
	\centering
	\subfloat[Images with `hotdog' label in confidence intervals]{
	    \label{fig:intro-img}
        \includegraphics[height=1.4in]{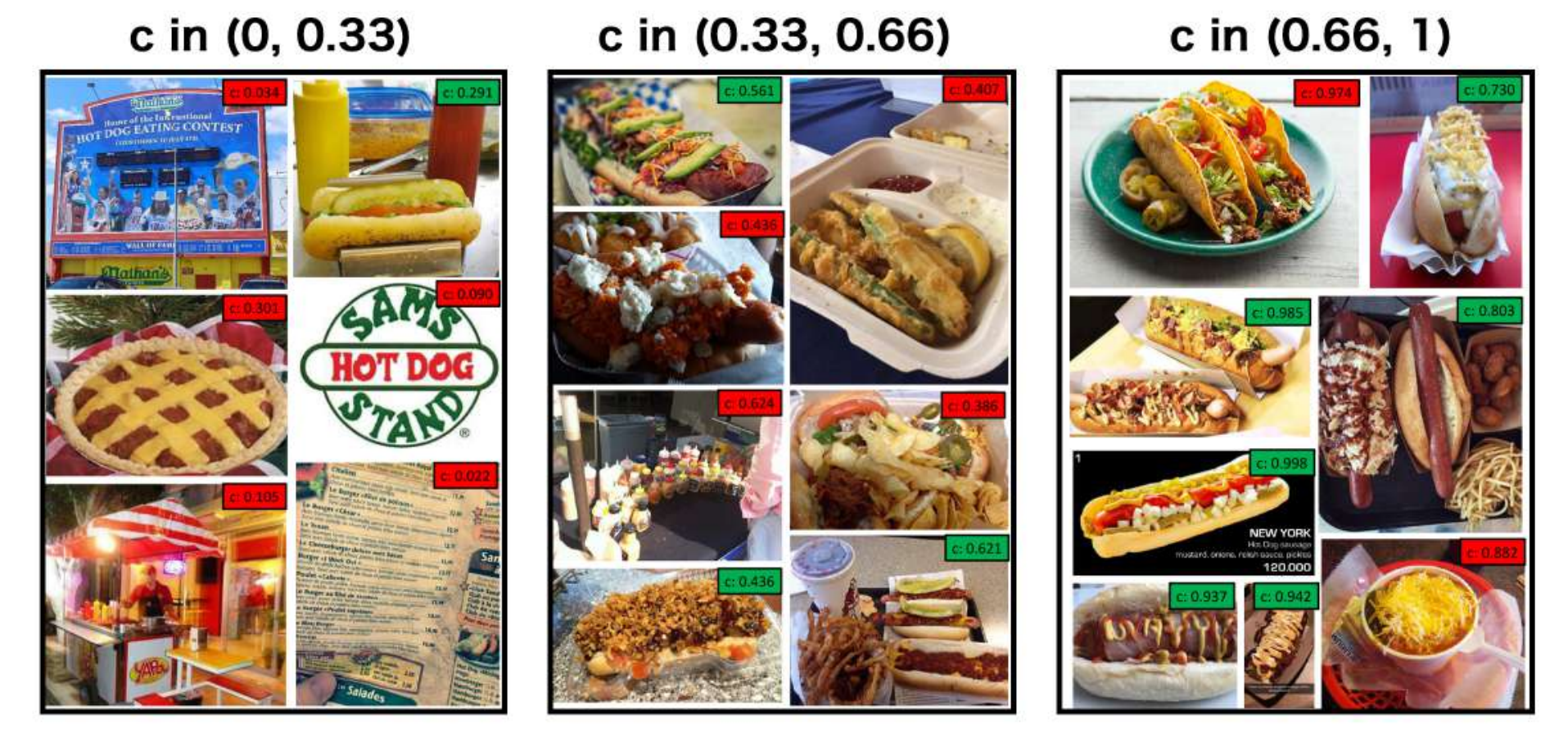}  
     }
     \hfill
     \subfloat[ECE Plot]{%
        \label{fig:intro-ece}
        \includegraphics[height=1.4in]{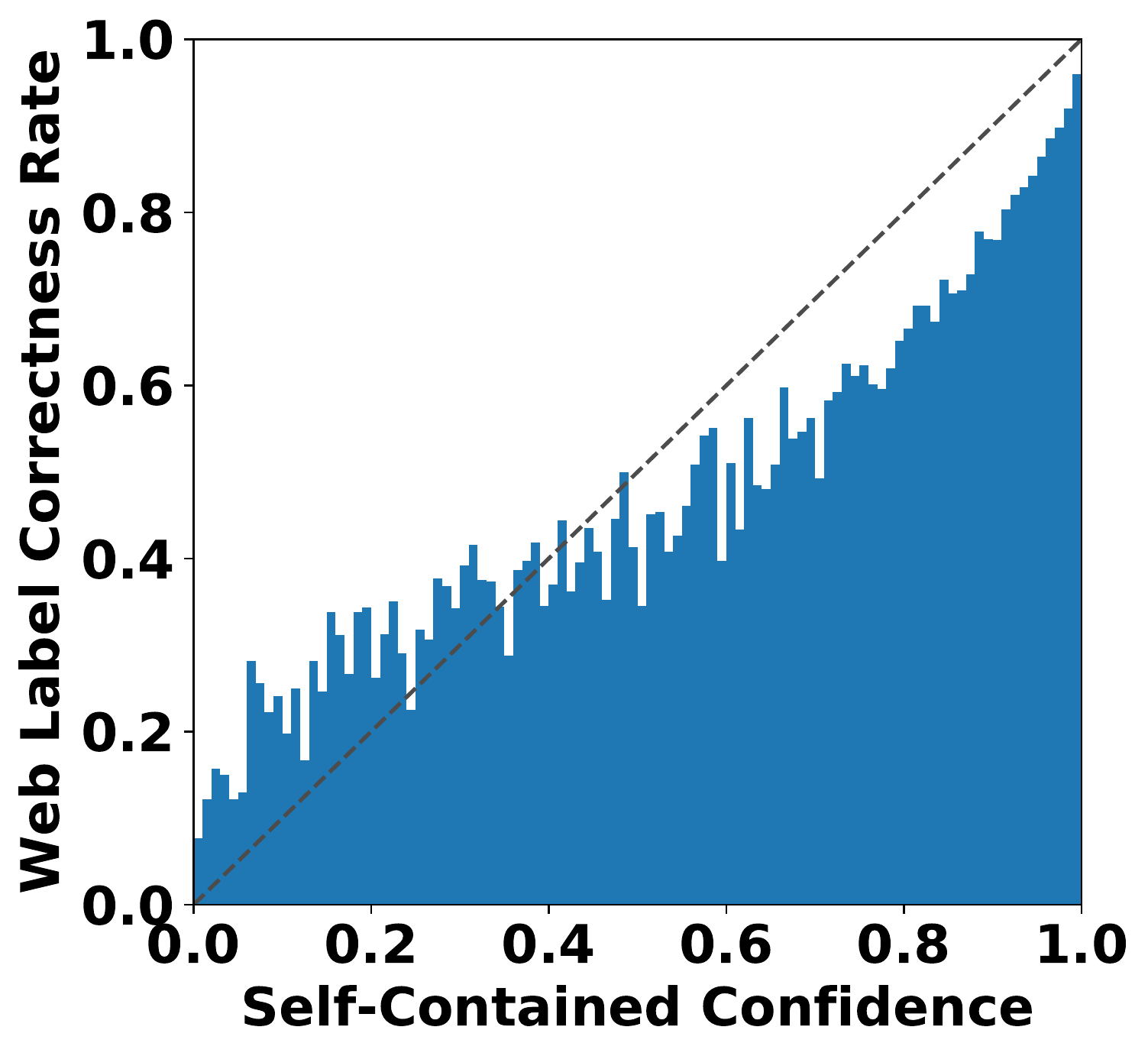} 
     }
	\caption{Exemplary images and ECE plot showing self-contained confidence (SCC) generally reflects web label correctness. A standard ResNet-50 is pretrained on the Food-101N training set for SCC extraction. (a) shows image samples grouped by low/medium/high confidences. The upper right tag on each image shows SCC value and the tag color indicates web label correctness (red: wrong, green: correct). (b) is the ECE plot using Food-101N human-verification set ($M=100$)}
	\label{fig:into}
\end{figure}

A straight-forward approach of WSL is to treat web labels as ground truth and all web samples are directly used to train DNNs~\cite{mahajan2018exploring,sun2017revisiting}. 
Some previous methods~\cite{jiang2018mentornet,lee2018cleannet} require additional clean subsets to learn a guidance model to judge the correctness of web labels and adopt a sample reweighting strategy for robust training of DNNs.
CurriculumNet~\cite{guo2018curriculumnet} avoids extra clean set by leveraging density assumption that samples from the high-density region are more reliable, and trains the model in a curriculum learning manner.
As all the above works only use webly supervised loss $\mathcal{L}_w$, recent works attempt to combine self-label supervised loss $\mathcal{L}_s$ with $\mathcal{L}_w$~\cite{han2019deep,tanaka2018joint}. $\mathcal{L}_s$ comes from the predictions of the model itself in a fashion of self-distillation~\cite{hinton2015distilling} or prototype-based rectification~\cite{snell2017prototypical}. 

Although it is promising to utilize $\mathcal{L}_s$ together with $\mathcal{L}_w$, we argue that the ratio balancing $\mathcal{L}_w$ and $\mathcal{L}_s$ should not be a constant across the entire dataset as in previous works~\cite{han2019deep,tanaka2018joint}. The correctness of web labels varies on a case-by-case basis, due to various causes of real-world label noise. Motivated by this observation, we design a framework that adaptively balances $\mathcal{L}_w$ and $\mathcal{L}_s$ on sample level.  

Inspired by the uncertainty prediction ability of DNNs~\cite{gal2016uncertainty}, we use DNN's prediction confidence, termed as self-contained confidence~(SCC), to achieve a sample-wise balance between $\mathcal{L}_w$ and $\mathcal{L}_s$. 
Model uncertainty shows how unsure the model considers its correctness on its own prediction, which is revealed by DNN's soft label output. 
When the model is trained with binary cross entropy~(BCE) loss, the model uncertainty can be estimated independently across all categories.
Here, we regard model uncertainty corresponding to the category of the sample's web label as SCC, reflecting the likelihood of web label correctness from the model's scope~\cite{gal2016uncertainty}.
Fig.~\ref{fig:intro-img} vividly shows a strong positive correlation between SCC and the correctness of web labels. This association is further confirmed by Expected Calibrated Error (ECE) plot~\cite{guo2017calibration}, who groups samples with SCC scores within an interval and calculates their average web label correctness rate using a human-annotated verification set from Food-101N~\cite{lee2018cleannet}. According to Fig.~\ref{fig:intro-ece}, samples who lie in higher SCC intervals generally have larger probabilities of correct web labels.

With SCC as an effective indicator of web label correctness, a generic SCC-based WSL framework is proposed. Intuitively, with the help of SCC, our framework enforces a webly supervised loss $\mathcal{L}_w$ if a web label is considered reliable, and a self-label supervised loss $\mathcal{L}_s$ otherwise. The self-label supervised loss utilizes the soft label predicted by a model pretrained on the WSL dataset as a self-supervised target. SCC, which is also extracted from the pretrained model, balances the ratio between $\mathcal{L}_w$ and $\mathcal{L}_s$ for each web sample. Our SCC is emphasized as `self-contained', as no extra guidance model or labeled clean dataset is needed. 
Following the uncertainty calibration approaches~\cite{guo2017calibration,thulasidasan2019mixup}, we also investigate the relationship between statistical metrics (e.g. ECE metric) and image classification accuracy.

Our contributions are summarized as follows:
\begin{itemize}
	\item A generic noise-robust WSL framework that does not require a human-verified clean dataset is proposed, novelly featured by sample-level confidence from the perspective of model uncertainty.
	\item Based on our framework, we further design a graph-enhanced mixup method that stands out among a series of SCC-friendly regularization methods to achieve better classification performance.
    \item We empirically conclude that under our framework, the statistical metrics of SCC are positively correlated with final classification accuracy, and self-label supervision is superior to consistency regularization for WSL tasks.
	\item The proposed framework achieves state-of-the-art results on two large-scale realistic WSL datasets,  WebVision-1000 and Food-101N.
\end{itemize}

\section{Related Work}

\subsection{Webly Supervised Learning}
Learning with noisy labels can be divided into two categories of problems according to sources of label noise, i.e., synthetic or realistic. For synthetic label noise, some works estimate a noisy channel (e.g., a transition matrix) to model the label noise~\cite{patrini2017making,xia2019anchor,yu2018efficient}. However, the designed or estimated channels might not stay effective in the real-world scenario.
WSL lies in the realistic noisy label problem. Seminal WSL works attempted to leverage a subset of human-verified samples, referred as `clean set'. 
MentorNet~\cite{jiang2018mentornet} learns a dynamic curriculum from the clean set for the sample-reweighting scheme, making the StudentNet only focus on probably correct samples. CleanNet~\cite{lee2018cleannet} transfers knowledge of label noise learned from a clean set with partial categories towards all categories, and adjust sample weights accordingly to alleviate the impact of noisy labels. In contrast to `clean set' prior, CurriculumNet~\cite{guo2018curriculumnet} assumes that samples with correct labels usually locate at high-density regions in visual feature space and designs a three-stage training strategy to train the model with data stratified by cleanness-levels.

Self labeling is another solution to purify noisy labels by replacing unreliable web labels with predictions by a model. Joint Optimization~\cite{tanaka2018joint} uses DNN's predictions as self labels, and Self-Learning~\cite{han2019deep} generates self labels by prototype voting and combines web labels and pseudo-labels using constant ratio. Compared to them, we balance self labels and web labels using sample-wise confidence, which relies on our observation that DNNs are capable of perceiving noisy labels with self-contained confidences. Self labels and confidences are unified in a single pretrained model in our approach. Table~\ref{tab:webly} clarifies the differences between other WSL solutions and ours.

\begin{table}[t]
    \begin{center}
	\caption{Highlighting the principal differences between other WSL methods and ours}
	\label{tab:webly}
		\begin{tabular}{@{\hskip 6pt}l@{\hskip 0pt}c@{\hskip 6pt}l@{\hskip 6pt}l@{\hskip 6pt}}
			\toprule
			{\multirow{2}{*}{Method}} & \multicolumn{1}{@{\hskip -10pt}l@{\hskip 6pt}}{Clean} &  \multicolumn{1}{l@{\hskip 6pt}}{Prior} & \multirow{2}{*}{How to suppress label noise?}  \\
			\multicolumn{1}{c}{} & \multicolumn{1}{@{\hskip -10pt}l@{\hskip 6pt}}{Set?} &  \multicolumn{1}{l@{\hskip 6pt}}{Knowledge} & \multicolumn{1}{@{\hskip 6pt}c}{}  \\
			\midrule
			MentorNet \cite{jiang2018mentornet}	  	  & \cmark   & Clean Set & Low weight on $\mathcal{L}_w$ for noisy samples  \\ 
			CleanNet \cite{lee2018cleannet}  		  	& \cmark   & Clean Set & Low weight on $\mathcal{L}_w$ for noisy samples \\ 
			CurriculumnNet \cite{guo2018curriculumnet}	& \xmark & Density &  Schedule noisy samples to later stages \\
			Joint Optim. \cite{tanaka2018joint}	& \xmark & Self-training &  Replace $\mathcal{L}_w$ with $\mathcal{L}_s$  \\
			Self-Learning \cite{han2019deep}	& \xmark & Density &  Combine $\mathcal{L}_w$ and $\mathcal{L}_s$ with constant-ratio \\
			\midrule
			Ours	  			   & \xmark & Uncertainty & Balance $\mathcal{L}_w$ and $\mathcal{L}_s$ sample-wisely\\ 
			\bottomrule
			\noalign{\bigskip}
		\end{tabular}
	\end{center}
\end{table}
	
\subsection{Semi-Supervised Learning}
Semi-supervised learning (SSL) utilizes a small fraction of labeled data and a large unlabeled data set altogether~\cite{zhu2005semi}. 
Solutions to SSL are basically within two main categories. 
One uses consistency regularization to ensure the model robustness by forcing networks producing identical predictions upon inputs with different augmentations, which is used in MixMatch \cite{berthelot2019mixmatch} and UDA \cite{xie2019unsupervised}. 
Another uses pseudo-labeling in the representative methods of Billion Scale~\cite{yalniz2019billion} and data distillation~\cite{radosavovic2018data}, which firstly trains models on the clean labeled set and then provides pseudo-labels for unlabeled data. 

The differences between WSL and SSL settings lead to key differences between our method and SSL methods. First, the self-label supervision in our method has a close connection with pseudo-labeling. However, our method utilizes all samples with both web labels and self labels, and SSL methods utilize a subset of unlabeled data with pseudo-labels only. The model for self-labeling in our method is learned from the entire noisy dataset, and the model for pseudo-labeling in SSL methods is trained on a small `clean' labeled set. Second, our self-label supervised loss has a similar form to consistency regularization. However, consistency regularization may be less powerful to correct the bias caused by label noise than cleaning the labels with self-labeling explicitly.
Details are discussed in Sec.~\ref{S:method-loss}.
	
\subsection{Model Uncertainty}
Model uncertainty refers to the level of distrust that the model considers its own prediction, which is vital for real-world applications. For classification tasks, the calculation is as simple as leveraging the highest score of the softmax output. To quantify the quality of model uncertainty, expected calibration error (ECE) is one widely used metric that claims an accurate uncertainty should align model predictions with classification accuracy~\cite{guo2017calibration,ovadia2019can}. For instance, if a network predicts a group of samples with a probability of $0.6$, we expect exactly $60\%$ samples of this group are classified correctly.

Following this path, several methods were proposed to improve the quality of uncertainty. Post-hoc calibration such as temperature scaling is one family of methods, which optimizes the mapping of produced uncertainty on the verification set~\cite{guo2017calibration}. However, such data-dependent rescaling methods cannot improve confidence quality fundamentally. Some other works explored within-training strategies that can provide high-quality model uncertainty, such as label smoothing~\cite{muller2019does}, dropout~\cite{gal2016dropout}, mixup~\cite{thulasidasan2019mixup}, Bayesian models~\cite{lakshminarayanan2017simple}, etc. AugMix~\cite{hendrycks2020augmix} is directly designed to improve uncertainty estimates through a data augmentation approach. However, few research works utilized the model confidence to architect model training.

In our work, model uncertainty is adapted for web label confidence estimation. Instead of using the maximum of the model's output probabilities, we pick the value on the exact web label from the probability distribution, which estimates the correctness of the sample's web label.\footnote{Web label confidence and self-contained confidence are used interchangeably throughout the paper.} Metrics such as ECE can also be adapted, i.e., web label confidence is considered well-calibrated if a model predicts all samples in a group with web label confidences of $0.6$, $60\%$ samples in this group have correct web labels. Being aware that the extraction of web label confidence requires the probability of each class to be calculated independently, binary cross-entropy (BCE) rather than softmax cross-entropy loss is used for training the network. 


\section{Proposed Method}
\begin{figure}[t]
	\centering
	\includegraphics[width=10cm]{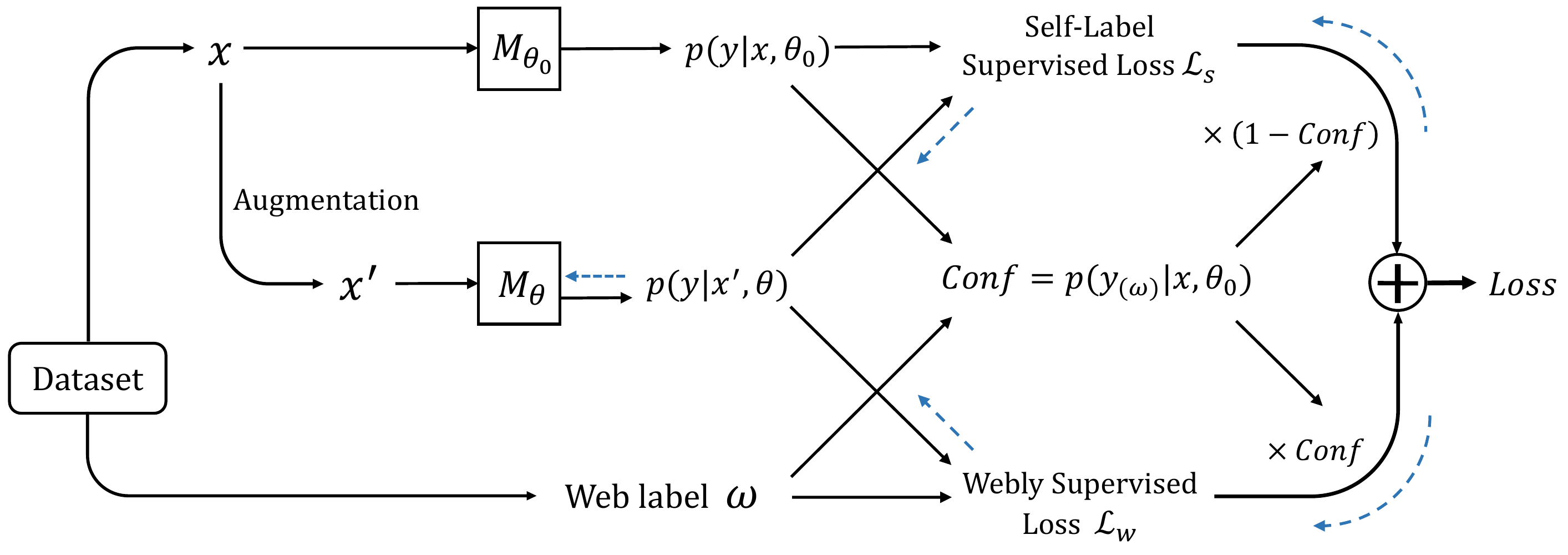}
	\smallskip
	\caption{Diagram of the proposed framework. Backward gradients pass through dashed arrows to update the only trainable parameter $\theta$. Pretrained model $M_{\theta_0}$ learns from entire WSL dataset to provide self label and SCC. $M_{\theta}$ is initialized by $M_{\theta_0}$}
	\label{Fig:pipeline}
\end{figure}

In this section, after a formal description of WSL task, we introduce two loss functions and the proposed framework with highlighted SCC. Our framework is compatible with various regularization methods. Especially, we propose graph-based aggregation (GBA) to enhance SCC for network training. The diagram of our framework is shown in Fig.~\ref{Fig:pipeline}.
	
\subsection{Webly Supervised Learning: Problem Statement and Notations}
\label{S:method-problem}
Webly Supervised Learning (WSL) aims at training an optimal deep neural network $\mathcal{M_\theta}$ from a dataset $\mathcal{D}=\{(x_1, y^*_1),\dots,(x_N, y^*_N)\}$ collected from the Internet.
$x_i$ denotes the $i$-th sample in the dataset, and the one-hot web label $y^*_i$ is the one-hot encoding of the web label $\omega_i$ (referring to $\omega_i$-th category). The web label $\omega_i$ is obtained from the search query of crawling the image $x_i$. Consider the massive noise in retrieved images from a search engine, $\omega_i$ or $y_i^*$ might not reflect the correct category that $x_i$ belongs to. Therefore, suppressing the noise in unreliable web labels becomes the main challenge in WSL.

For convenience, we use symbols $x$, $y^*$, $\omega$ directly to represent an arbitrary sample, its one-hot web label and its web label, respectively. For the multi-label problem, $y_{(j)}$ denotes sample's label on $j$-th class.
$p(y|x,\theta)$ denotes the label prediction of sample $x$ by the model $\mathcal{M}_\theta$.
	
\subsection{Webly Supervised Loss and Self-Label Supervised Loss}
\label{S:method-loss}
Webly supervised loss and self-label supervised loss are two widely adopted loss functions in WSL~\cite{guo2018curriculumnet,han2019deep,reed2014bootstrap,tanaka2018joint}. 
Webly supervised loss utilizes web labels as supervision information, and self-label supervised loss~\cite{han2019deep,tanaka2018joint} utilizes predictions of a pretrained model instead. Formally, we define them as follows.

For webly supervised loss, given $x'$ augmented from $x$ with web label $\omega$, the loss function can be expressed as
\begin{equation}
\label{eq:loss_w}
	\mathcal{L}_w=-\biggl[\log\left(p(y_{(\omega)}|x',\theta)\right)
	+\sum\nolimits_{j\in\mathcal{S}\backslash\omega}\log\left(1-p(y_{(j)}|x',\theta)\right)\biggl].
\end{equation}
Notice that webly supervised loss is in the form of binary cross-entropy (BCE) loss, because a webly-crawled image probably has multi-label semantics.

For self-label supervised loss, we use the prediction of the pretrained model $\mathcal{M}_{\theta_0}$, which is trained directly on the original web label dataset. As the predictions on samples will be used for finetuning $\mathcal{M}_{\theta_0}$ itself, We call them self labels.  Therefore, with self label $p(y|x,\theta_0)$ from model $\mathcal{M}_{\theta_0}$, the self-label supervised loss is 
\begin{equation}
\mathcal{L}_s = -\sum_{j\in\mathcal{S}}
\biggl[p(y_{(j)}|x,\theta_0)\log\left(p(y_{(j)}|x',\theta)\right) 
+ \left(1-p(y_{(j)}|x,\theta_0)\right)\log\left(1-p(y_{(j)}|x',\theta)\right)
\biggl],
\end{equation}
where $y_{(j)}$ represents the prediction for $j$-th class in label set $\mathcal{S}$ for multi-class classification problem.

A similar loss to self-label supervised loss is consistency loss~\cite{berthelot2019mixmatch,xie2019unsupervised}, which provides an auxiliary regularization by enforcing a model to output similar predictions on different augmented counterparts of the same image. Consistency loss is proven to be effective on a large number of unlabeled images for semi-supervised learning. In WSL, however, as the quality of self labels can be guaranteed by feeding a pretrained model with weak augmented images, we found that the high-quality self-supervised loss is more effective than auxiliary consistency loss. An experimental comparison will be shown in Sec.~\ref{S:exp-ablation}.

\subsection{Self-Contained Confidence}
\label{S:method-conf}
It is desirable to adaptively balance webly supervised loss and self-label supervised loss on sample level. Intuitively, we should trust webly supervised loss more on samples with reliable web labels, while self-label supervised loss would dominate the total loss confronting incorrect web labels. 

In our method, model $\mathcal{M}_{\theta_0}$ provides only self labels, but also the reliability of web labels. Notice that with BCE loss, model $\mathcal{M}_{\theta_0}$ predicts the probability that $x$ belongs to class $i$ as $p(y_{(i)}|x,\theta_0)$. Specially, we focus on the model prediction on the one-hot web label $y^*$ whose category index is $\omega$, denoted as $p(y_{(\omega)}|x, \theta_0)$. 
Therefore, the only trainable parameter $\theta$ would be updated by minimizing the final loss 
\begin{equation}
	\mathcal{L}=c\times\mathcal{L}_w+(1-c)\times\mathcal{L}_s, \quad \text{where}~ c=p(y_{(\omega)}|x, \theta_0).
\end{equation}
The confidence $c$ is named as \emph{self-contained confidence} (SCC), as it is self contained in the pretrained model and requires no extra data or knowledge.

\subsection{Graph-Based Aggregation}
\label{S:method-model}
A key component in the proposed method is the pretrained model $\mathcal{M}_{\theta_0}$ for estimating both SCC and self labels. As the model is trained on noisy web labels, we employ
mixup~\cite{zhang2018mixup}, which is known as an effective regularization to make DNNs less prone to over-confident predictions and predicted scores of DNNs better calibrated to the actual confidence of a correct prediction~\cite{thulasidasan2019mixup}.

In addition, we propose a graph-based aggregation (GBA) method to further boost the confidence quality and classification performance. GBA does a smoothing operation on a visual similarity graph spanned by image features. By viewing every image as a node, a $k$-nearest-neighbor ($k$-NN) graph is firstly constructed based on features located before \textit{fc} layer of pretrained model $\mathcal{M}_{\theta_0}$. Cosine similarity of features is computed across every pair in the neighborhood as edge weight. Hereby, an undirected $k$-NN graph with weighted adjacent matrix $\mathbf{A}$ is obtained. Let $\mathbf{P}$ denote a matrix of self labels, and the corrected self labels after GBA are denoted as
\begin{equation}
\mathbf{\hat{P}}=\mathbf{D}^{-\frac{1}{2}}  \left(\lambda\mathbf{I}+\mathbf{A}\right)   \mathbf{D}^{-\frac{1}{2}}\mathbf{P},
\end{equation}
where $D(i,i)=\lambda+\sum_{j=1}^{N}{A(i,j)}$. $\lambda$ controls the portion of original self labels in the post-GBA self labels. SCC will also be extracted from $\mathbf{\hat{P}}$.
GBA is a post-processing step with graph filtering~\cite{kipf2016gcn} and complementary to other methods such as mixup. We evaluate several potential methods and conclude mixup + GBA leads to the optimal performance in Sec. \ref{S:exp-regular}. 
	
\section{Experiments}
In this section, we firstly introduce three public WSL datasets. Then, we investigate several SCC-friendly methods, among which GBA-enhanced mixup stands out as the best one in both statistical metrics and classification accuracy. More ablation studies demonstrate the effectiveness of both sample-wise adaptive loss and self-label supervision. Finally, we show that the proposed method reaches the state-of-the-art on the public WSL datasets. We leave the exploration of robustness of our framework and formal algorithm in the Appendix.

\subsection{Datasets and Configurations}
\label{S:Dataset}
\textbf{WebVision-1000}~\cite{li2017WebVision} contains $2.4M$ noisy-labeled training images crawled from Flickr and Google, with keywords from $1000$ class-labels in ILSVRC-2012~\cite{deng2009imagenet}.
The estimated web label accuracy is $48\%$~\cite{guo2018curriculumnet}. 
The ILSVRC-2012 validation set is also utilized along with WebVision-1000's own validation set.

\textbf{WebVision-500} is a quarter-sized version of WebVision-1000 for evaluation and ablation study in low cost without losing generalization. We randomly sample one-half categories with one-half samples in the training set, and keep the full validation set of the selected $500$ categories. This dataset is used for our ablation study in Sec.~\ref{S:exp-regular},~\ref{S:exp-conf},~\ref{S:exp-ablation}. 

\textbf{Food-101N}~\cite{lee2018cleannet} is another web dataset with $310k$ images classified into $101$ food categories. Images are crawled from Google, Yelp, etc. We evaluate our model on the test set of Food-101~\cite{bossard2014food}, Food-101N's clean dataset counterpart. $60k$ human verification labels are provided, indicating the correctness of web labels. The estimated label accuracy is around $80\%$.

\textbf{Configuration details.}
ResNet50 is selected as our CNN model in all experiments~\cite{he2016deep}. 
For more efficient training on WebVision, a minor-revised ResNet50-D is utilized~\cite{he2019bag}. Food101N uses standard ResNet50 for a fair comparison.
We use the following settings that completely refer to~\cite{he2019bag}.
Batch size is set as 256 and mini-batch size as 32. We use the standard SGD with the momentum of $0.9$ and weight decay of $10^{-4}$. A warm-start linearly reaches the initial learning rate (LR) in the first $10$ epochs. The remained epochs are ruled by a cosine learning rate scheduler. A simple class reweighting is performed to deal with class imbalance. The initial LR is $0.1$ with total $L$ epochs for pretrained models. The main model has initial LR of $0.05$ with identical epoch numbers. $L$=120 for WebVision-500 and Food101N, $L$=150 for WebVision-1000.
	
\subsection{Exploring Optimal Regularization Method}
\label{S:exp-regular}
In this section, we experiment with seven different confidence-friendly regularization methods for $\mathcal{M}_\theta$ under our framework. We conclude that GBA-enhanced mixup (mixup+GBA) is the most efficient one for the best performance. However, as the main contribution of our work is the simple yet effective noise-robust pipeline with SCC, regularization is not a necessary part of our model.

    
Besides the standard setting with BCE loss, which is denoted as `\textbf{Vanilla}', we introduce the following  regularization methods for model $\mathcal{M}_\theta$.

\textbf{Label Smoothing} prevents over-confidence problems by adding a small value of $\epsilon$ on the zero-values in one-hot encoding labels \cite{szegedy2016rethinking}. We use $\epsilon=0.1$.

\textbf{Entropy Regularizer} discourages over-confident model prediction by adding a penalizing term to standard loss functions \cite{pereyra2017regularizing}. Regularizer weight is set as $0.1$.

\textbf{MC Dropout} is selected as the representation of Bayesian methods. It approximates Bayesian inference by randomness in dropout operation \cite{gal2016dropout}. Dropout rate $p$ is set $0.5$. When testing, we infer $50$ times and average the predictions.

\textbf{Mixup} is a simple but effective pre-processing method that convexly combines every pair of two sampled images and labels \cite{zhang2018mixup}. 
\cite{thulasidasan2019mixup} proves its strong uncertainty calibration capability beyond its label smoothing effects.

\textbf{AugMix} is another data augmentation method with consistency loss, which produces well-calibrated model uncertainty \cite{hendrycks2020augmix}.

\textbf{Ensemble} utilizes several models with identical tasks to boost the ultimate performance \cite{dietterich2000ensemble}. With $E$ vanilla models with different random initializations, we average their predictions on every sample.

\textbf{Graph-based Aggregation (GBA)} is introduced in Sec.~\ref{S:method-model}. We use $k=10$ and $\lambda=0.5$ as hyper-parameters.
	
\begin{table}[t]
	\caption{Performance of the pretrained model (S1) and finetuned model (S2)}
	\smallskip
	\centering
	\begin{tabular}{lcccccccc}
		\toprule
		{\multirow{2}{*}{Method}} & \multicolumn{2}{c}{S1-WebVision} & \multicolumn{2}{c}{S1-ImageNet} & \multicolumn{2}{c}{S2-WebVision} & \multicolumn{2}{c}{S2-ImageNet}\\
		\multicolumn{1}{c}{} & \multicolumn{1}{c}{Top-1} & \multicolumn{1}{c}{Top-5} & \multicolumn{1}{c}{Top-1} & \multicolumn{1}{c}{Top-5}  & \multicolumn{1}{c}{Top-1} & \multicolumn{1}{c}{Top-5} & \multicolumn{1}{c}{Top-1} & \multicolumn{1}{c}{Top-5}\\
		\midrule
		Vanilla 						& 75.42 & 88.65 & 68.84 & 84.62 & 76.46 & 89.63 & 69.78 & 85.32 \\ 
		Label Smoothing  		 & 75.81 & 89.32 & 69.11 & 85.54  & 77.02 & 90.33 & 70.86 & 86.71\\ 
		Entropy Regularizer 	& 74.77 & 89.44 & 68.80 & 85.81  & 73.78 & 88.76 & 67.99 & 85.36\\ 
		MC Dropout $(p=0.5)$ & 75.16 & 88.90 & 68.73 & 84.60 & 76.00 & 89.50 & 69.78 & 86.01 \\ 
		Mixup $(\alpha=0.2)$  & 76.35 & 90.31 & 71.15 & \textbf{87.36} & 77.47 & 91.02 & 72.25 & 88.47\\
		AugMix  						& 76.61 & 89.58 & 69.06 & 84.30 & 76.96 & 90.10 & 69.61 & 85.32\\
		Ensemble $(E=5)$ 	  & \textbf{78.98} & \textbf{91.27} & \textbf{72.45} & 87.26  & \textbf{79.12} & \textbf{91.73} & \textbf{72.73} & 87.96 \\ 
		\midrule
		Vanilla + GBA			   & 75.42 & 88.65 & 68.84 & 84.62 & 77.12 & 90.73 & 71.56 & 87.78\\
		Mixup + GBA			   & 76.35 & 90.31 & 71.15 & 87.36 & 77.76 & 91.43 & 72.59 & \textbf{88.65}\\
		\bottomrule
		\noalign{\bigskip}
	\end{tabular}
	\label{tab:stage1}
\end{table}

\textbf{Result Analysis}.
Table~\ref{tab:stage1} reports the results. S1 is short for the pretraining stage for $\mathcal{M}_{\theta_0}$, S2 for the finetuning stage using our framework. Generally, good performance in S1 favors S2. Mixup and Ensemble are the two most effective regularizers. As mentioned in Sec.~\ref{S:method-model}, Mixup smooths discriminative spaces and ensemble averages models’ biases. The advantages of these two methods are combined in GBA design, as a graph smoothing operator for neighbor predictions, which is proven effective empirically. Improvement from GBA is weaker on mixup compared to vanilla since mixup offers the same effect of smoothing space with GBA. However, mixup+GBA still reaches the optimal result besides the costly ensemble method.

\subsection{Understanding Self-Contained Confidence}
\label{S:exp-conf}
As SCC plays a critical role in our framework, we explore an interesting question: 
how great the SCC quality affects the final accuracy reported in the previous section? We also show the relationship between three statistical metrics adapted from uncertainty theories and our accuracy-based metric. 

For statistical metrics, we manually create a verification set $\mathcal{V}=\{v_1, \dots, v_n\}$ for WebVision-500 by annotating whether the web label is correct on $n=12500$ samples, with $50$ randomly sampled cases from $250$ random classes. 

To evaluate the quality of SCC, The following metrics are utilized.

\textbf{Second-stage Accuracy on Vanilla (SAV)}. 
To empirically evaluate different SCCs, we use an identical vanilla pretrained model for self-labeling and finetuning under our framework. Therefore, the accuracy of second-stage finetuned model is only determined by the quality of SCC. Note that the models for producing SCC are different and with different regularization methods.
	
\textbf{Mean Square Error (MSE)}. Verification set $\mathcal{V}$ can be considered as a set of ground-truth confidence since it values $1$ with the correct web label and values $0$ when incorrect. Thus, MSE estimates the squared difference between the given confidence and the ground-truth, which is defined as 
\begin{equation}
\label{E:mse}
\textit{MSE} = \frac{1}{n}\sum_{i=1}^{n}(v_i - c_i)^2.
\end{equation}

\textbf{Expected Calibration Error (ECE)}. Calibration error is originally used to evaluate the model interpretability on their predictions \cite{guo2017calibration}, while we slightly adapt it for confidence quality evaluation. 
Formally, in the verification set $\mathcal{V}$, for all samples whose confidences fall into $(\frac{m-1}{M},\frac{m}{M}]$ form the $m$-th bin, where average confidence $\textit{conf}(B_m)=\frac{1}{|B_m|}\sum_{i\in B_m}c_i$ and the average web-label reliability $\textit{rel}(B_m)=\frac{1}{|B_m|}\sum_{i\in B_m}v_i$ are calculated. Thus, ECE is defined as
\begin{equation}
\label{E:ece}
\textit{ECE} = \sum_{m=1}^{M}\frac{|B_m|}{n}\biggl|\textit{rel}(B_m) - \textit{conf}(B_m)\biggr|.
\end{equation}

\begin{table*}[t]
	\centering
	\caption{Evaluations of SCC provided by different methods. Column $1$-$3$ reports statistical metrics MSE, ECE and OCE. Column $4$-$7$ reports model-based metric SAV}
	\smallskip
	\begin{tabular}{lccc@{\hskip 5pt}cccc}
		\toprule
		\multirow{2}{*}{Confidence Provider} & \multicolumn{1}{c}{\multirow{2}{*}{MSE}} & \multicolumn{1}{c}{\multirow{2}{*}{ECE}} & \multicolumn{1}{c}{\multirow{2}{*}{OCE}} & \multicolumn{2}{c}{SAV-WebVision} & \multicolumn{2}{c}{SAV-ImageNet}\\
		& \multicolumn{1}{c}{} & \multicolumn{1}{c}{} & \multicolumn{1}{c}{} & \multicolumn{1}{c}{Top-1} & \multicolumn{1}{c}{Top-5} & \multicolumn{1}{c}{Top-1} & \multicolumn{1}{c}{Top-5}\\ 
		\midrule
		Vanilla 						& 0.2795 & 0.2371 & 0.1518 & 76.46 & 89.63 & 69.78 & 85.32 \\ 
		Label Smoothing  		 & 0.2786 & 0.2280 & 0.1200 & 76.82 & 89.86 & 70.06 & 85.76 \\ 
		Entropy Regularizer 	& 0.4137 & 0.4138 & 0.0370 & 76.40 & 90.17 & 70.31 & 86.36 \\ 
		MC Dropout $(p=0.5)$ & 0.2807 & 0.2431 & 0.1193 & 76.68 & 89.89 & 70.34 & 85.97 \\ 
		Mixup $(\alpha=0.2)$  & \textbf{0.2510} & \textbf{0.1828} & 0.0135 & 77.14 & 90.18 & \textbf{71.00} & 86.48 \\
		Augmix	  					& 0.2869 & 0.2366 & 0.1757  & 76.67 & 89.65 & 69.89 & 85.63  \\ 
		Ensemble $(E=5)$ 	  & 0.2687 & 0.2233 & 0.1537  & 76.49 & 89.68 & 70.06 & 85.86  \\ 
		Vanilla + GBA 	  & 0.2612 & 0.2494 & \textbf{0.0002}  & \textbf{77.17} & \textbf{90.55} & 70.89 & \textbf{86.84} \\ 
		\bottomrule
	\end{tabular}
	\label{tab:comparison}
\end{table*}

\begin{figure}[t!]
    \centering
    \subfloat[Vanilla]{
        \label{fig:sub-vanilla}
        \includegraphics[width=.201\textwidth]{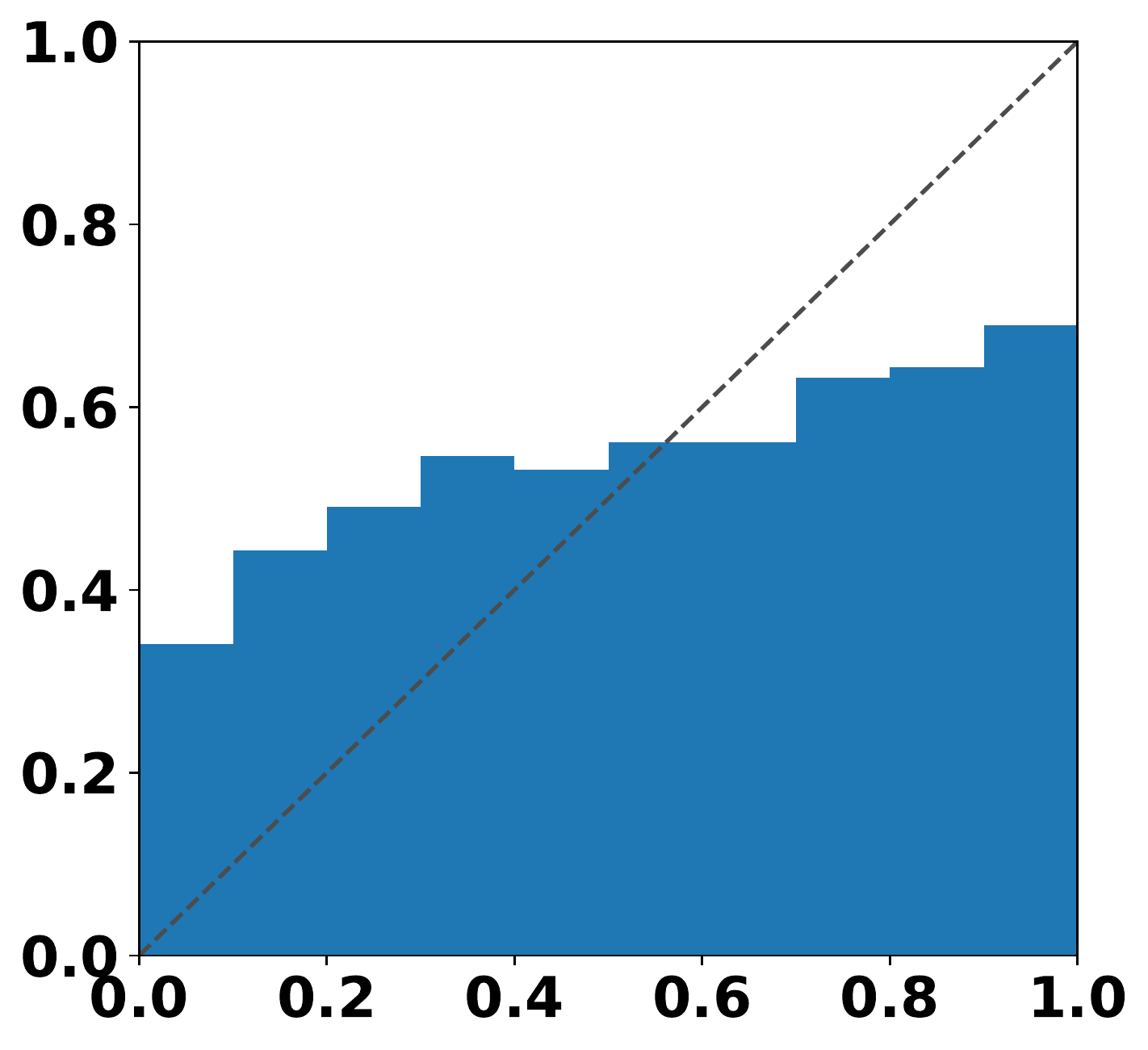}
    }
    \subfloat[Label Smooth]{
        \label{fig:sub-labelsmooth}
        \includegraphics[width=.201\textwidth]{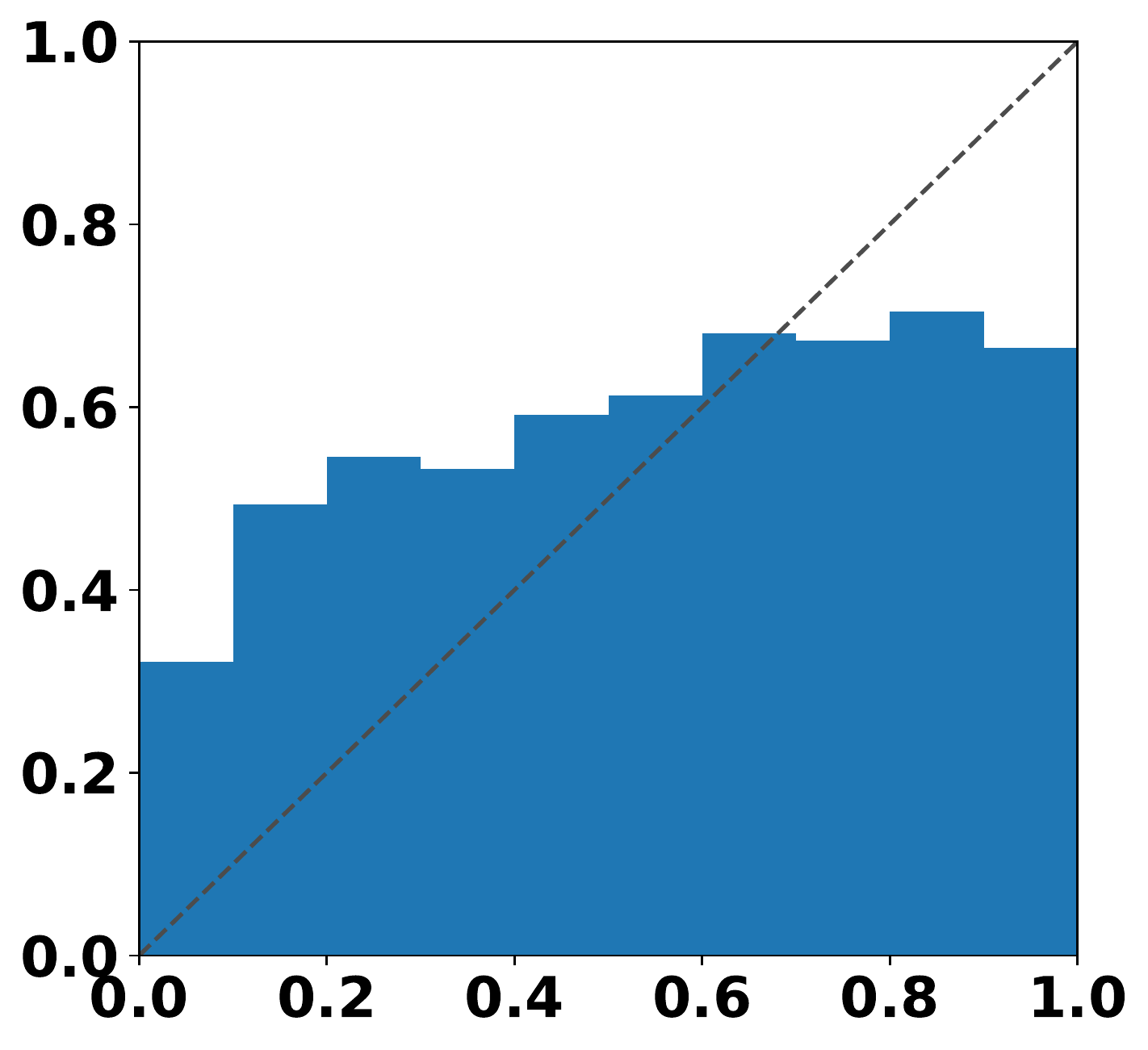}
    }
    \subfloat[Entropy Reg.]{
        \label{fig:sub-penalty}
        \includegraphics[width=.201\textwidth]{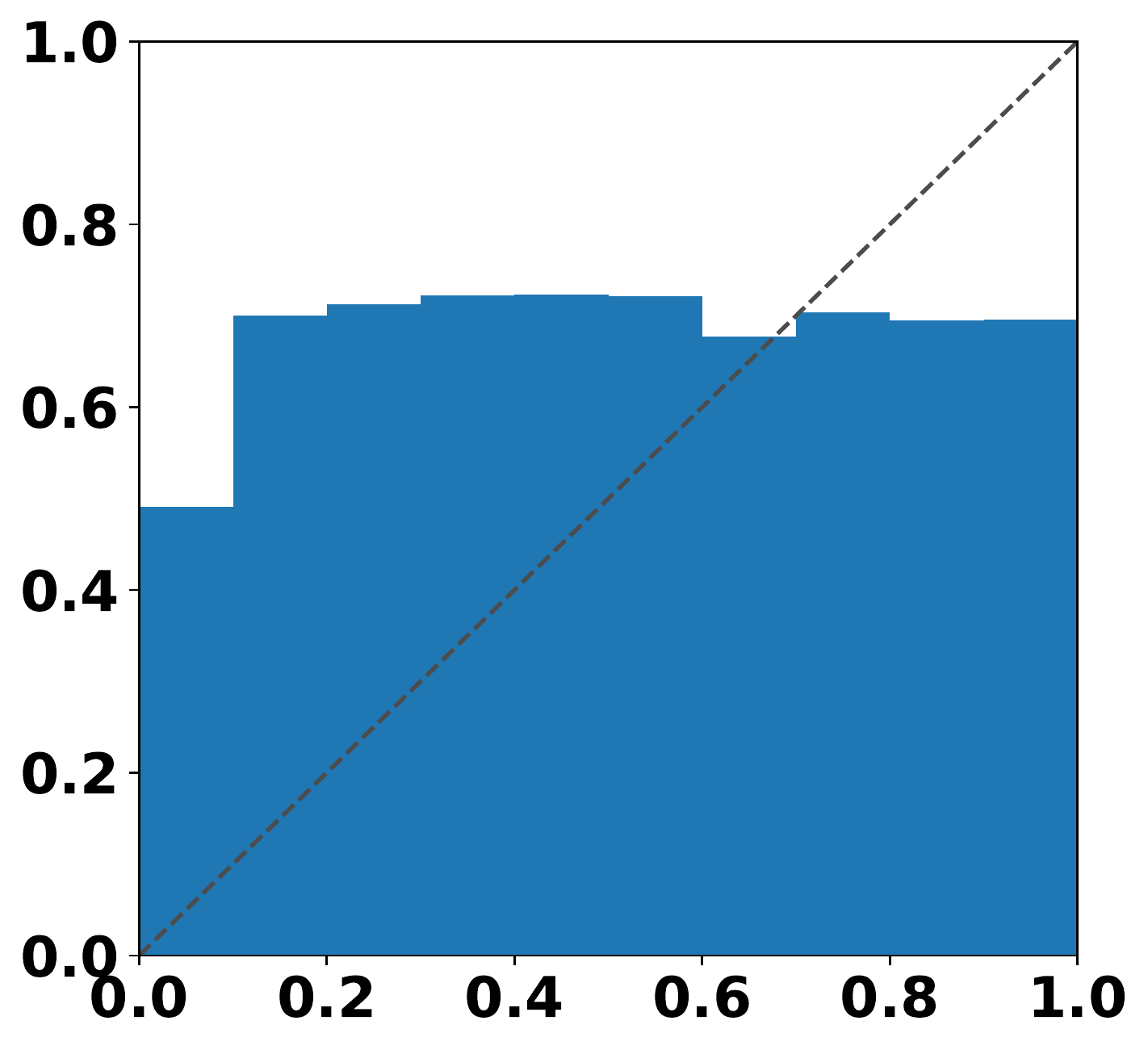}  
    }
    \subfloat[MC Dropout]{
        \label{fig:sub-dropout}
        \includegraphics[width=.201\textwidth]{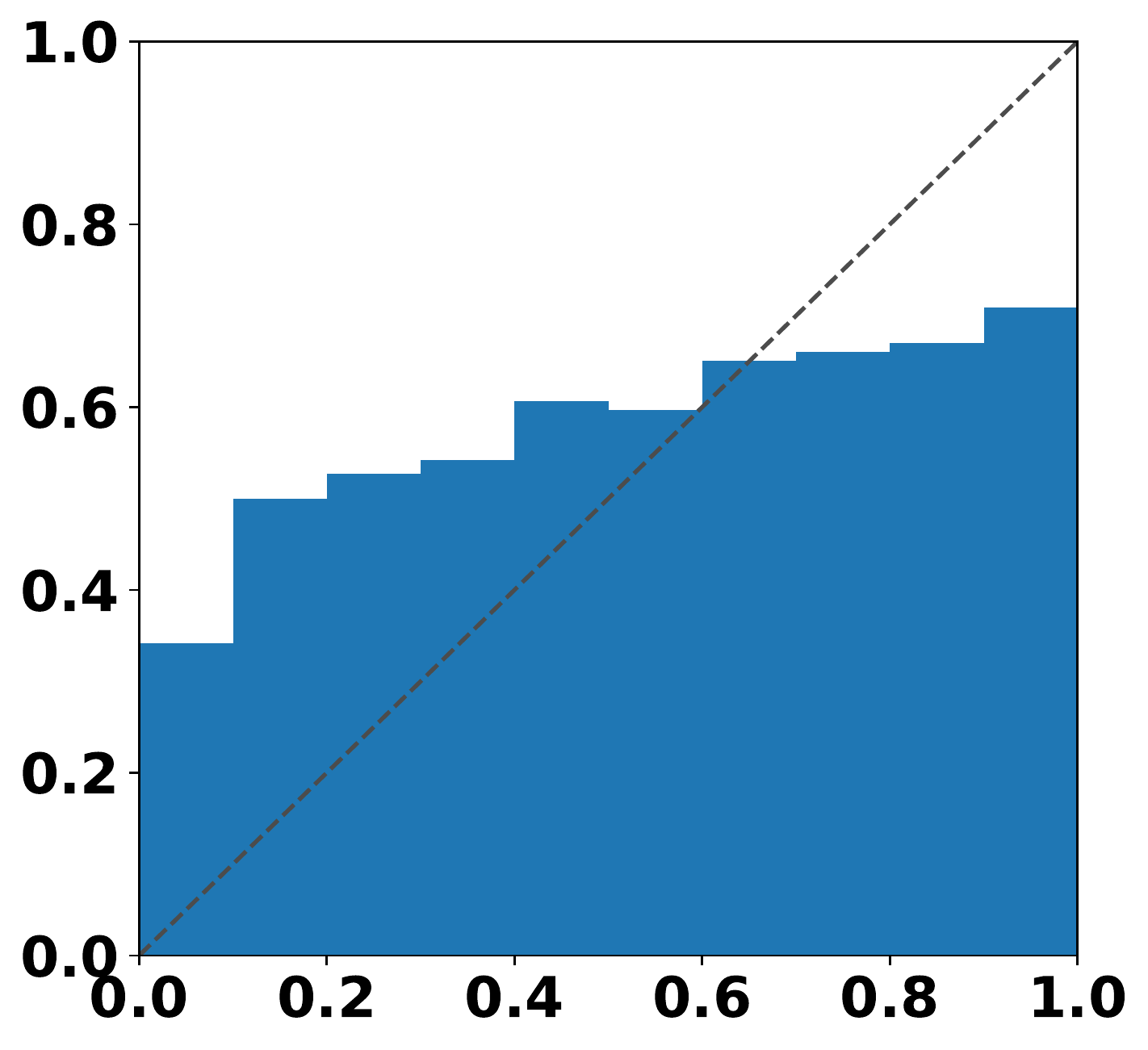}
    }
    
    \subfloat[Mixup]{
       	\includegraphics[width=.201\textwidth]{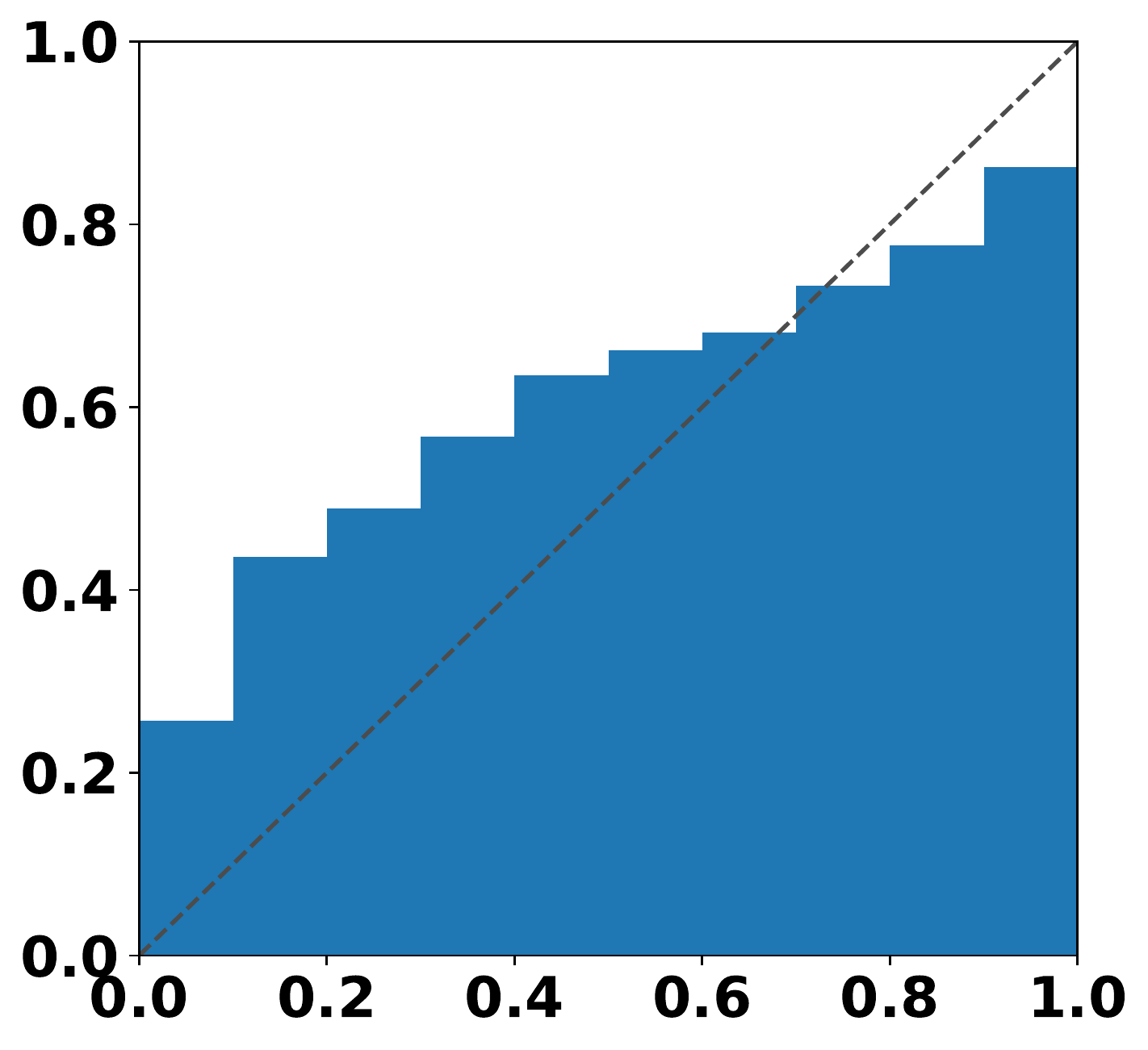}  
		\label{fig:sub-mixup}
    }
    \subfloat[AugMix]{
   		\includegraphics[width=.201\textwidth]{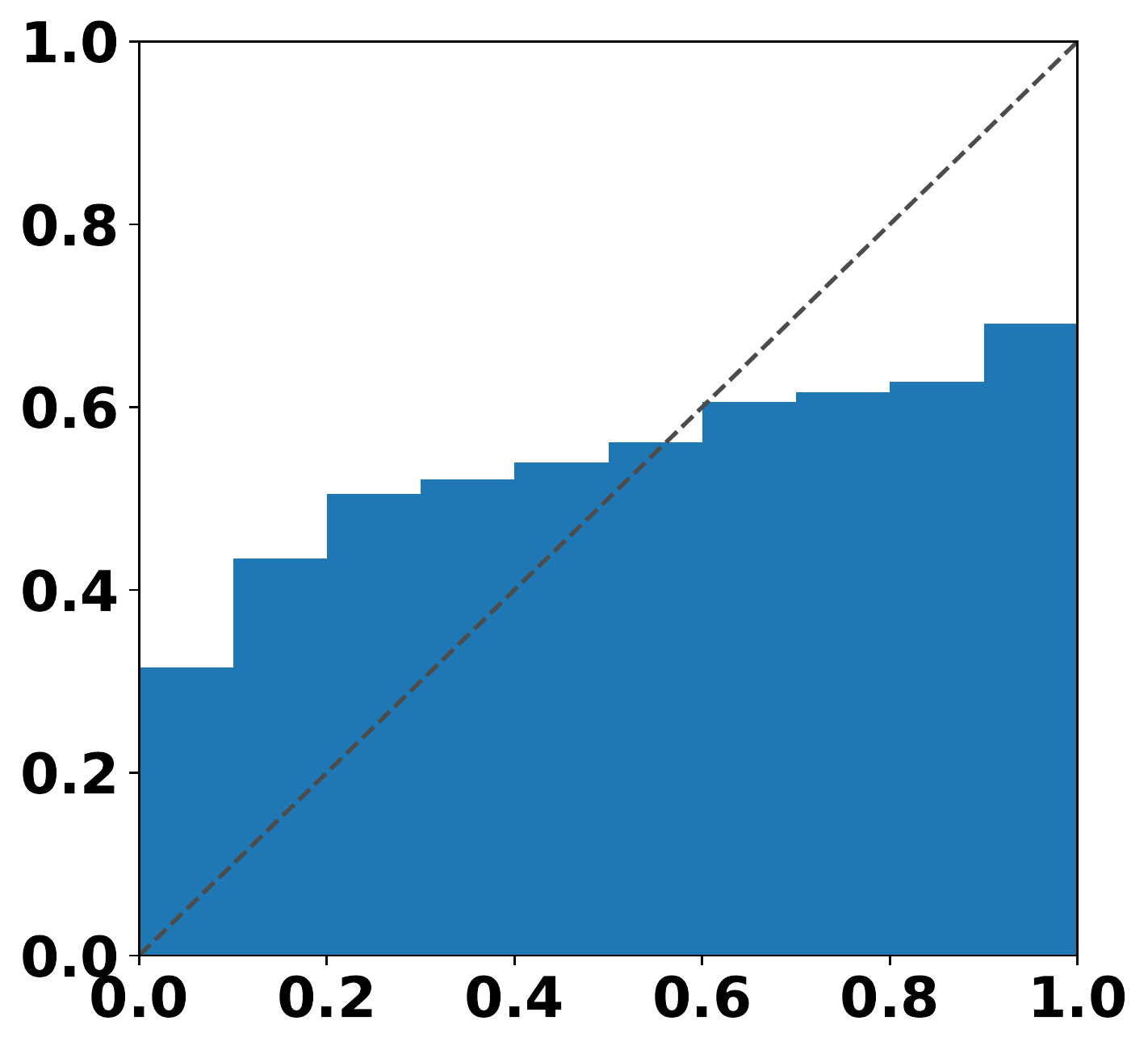}  
		\label{fig:sub-augmix}
    }
    \subfloat[Ensemble]{
        \includegraphics[width=.201\textwidth]{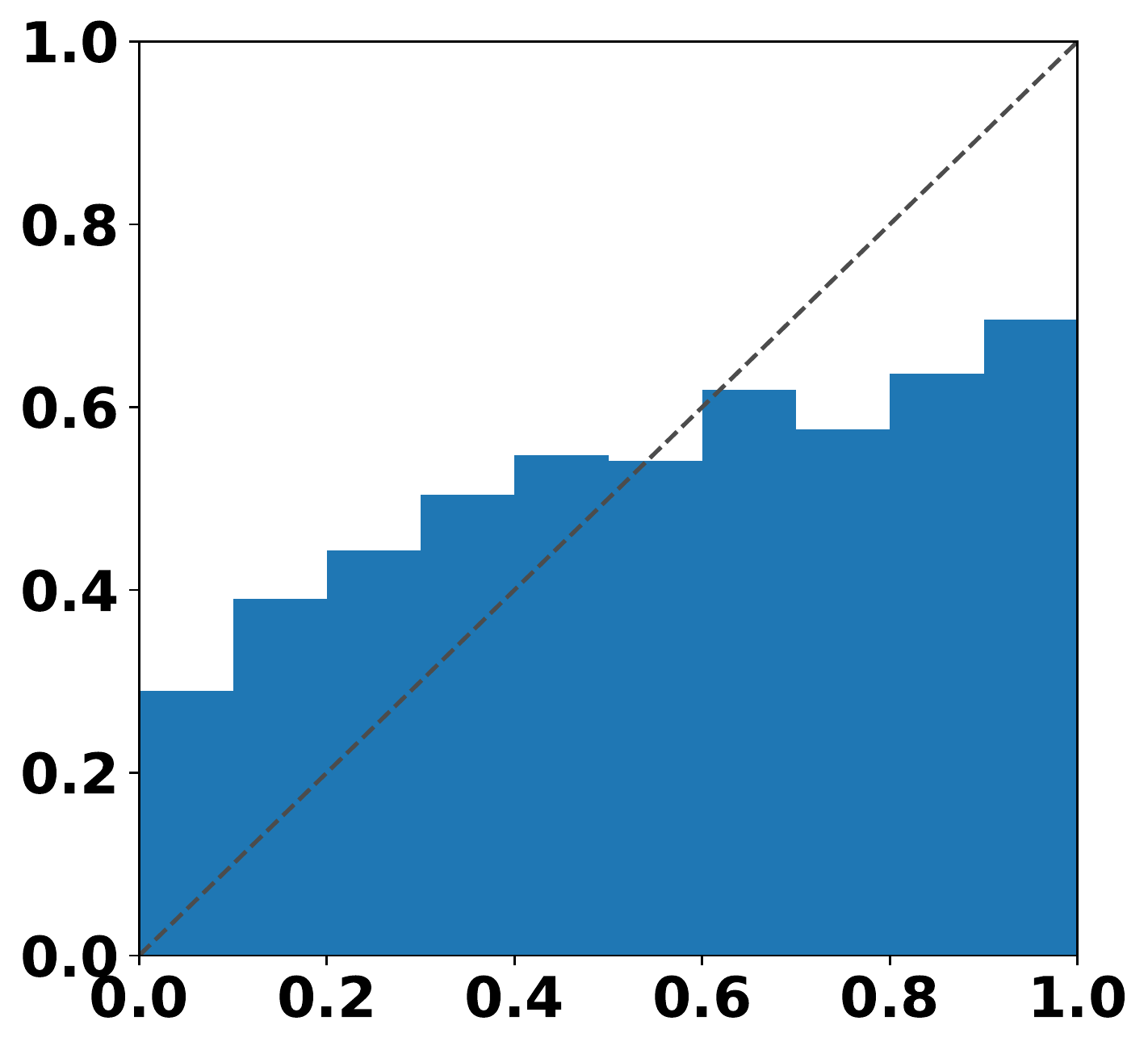}  
		\label{fig:sub-ensemble}
    }
    \subfloat[GBA]{
    	\includegraphics[width=.201\textwidth]{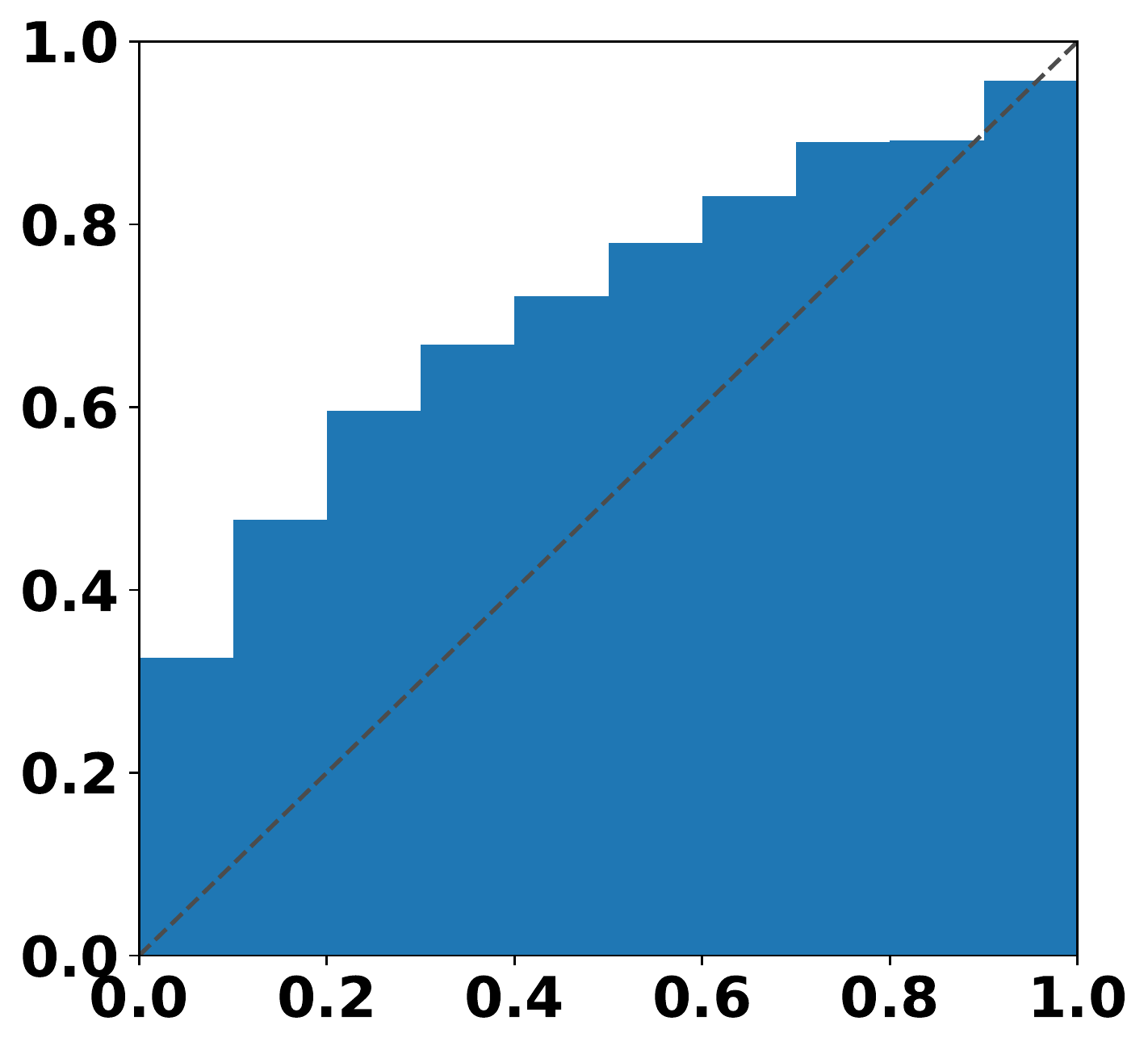}  
		\label{fig:sub-gba}
    }
    \caption{ECE diagrams of confidences from different SCC provider}
	\label{fig:ece}
\end{figure}
\textbf{Over-Confidence Error (OCE).} Samples with high SCC but incorrect web labels are especially harmful to our framework, since introducing the wrong web label is much worse than using self labels. OCE evaluates the level of over-confidence by punishing more on higher-confident bins with low reliability, defined as
\begin{equation}
\label{E:oce}
\textit{OCE} = \sum_{m=1}^{M}\frac{|B_m|}{n}\left[\textit{conf}(B_m)\times \max\Bigl\{\textit{conf}(B_m)-\textit{rel}(B_m), 0\Bigr\} \right].
\end{equation}
In this work, we calculate ECE and OCE with $M=100$. For visualization in Fig. \ref{fig:ece}, we use $M=10$. Fig.\ref{fig:intro-ece} uses $M=100$.

\textbf{Result Analysis}.
Best metric performance is reached by either mixup or Vanilla+GBA, while the ensemble also produces a good result. Fig.~\ref{fig:ece} visualizes ECE diagrams, where GBA and mixup look more calibrated than any other model. A similar result is shown in Tabel~\ref{tab:comparison} Column $2$-$4$. Column $5$-$8$ presents the metric of SAV which shows GBA provides good quality confidence that favors our proposed framework. According to our exploration of SCC, we conclude the following insights:
(1) SCC can reflect the reliability of the web label according to Fig.~\ref{fig:ece};
(2) SCC plays a key role in our pipeline through adaptively balancing two losses on the sample level since empirical metric SAV is generally proportional to the statistical metric ECE.
    
\subsection{Ablation Study}
\label{S:exp-ablation}

\begin{figure}[b]
\centering
\subfloat[Sample-wise confidence]{
    \includegraphics[width=.43\textwidth]{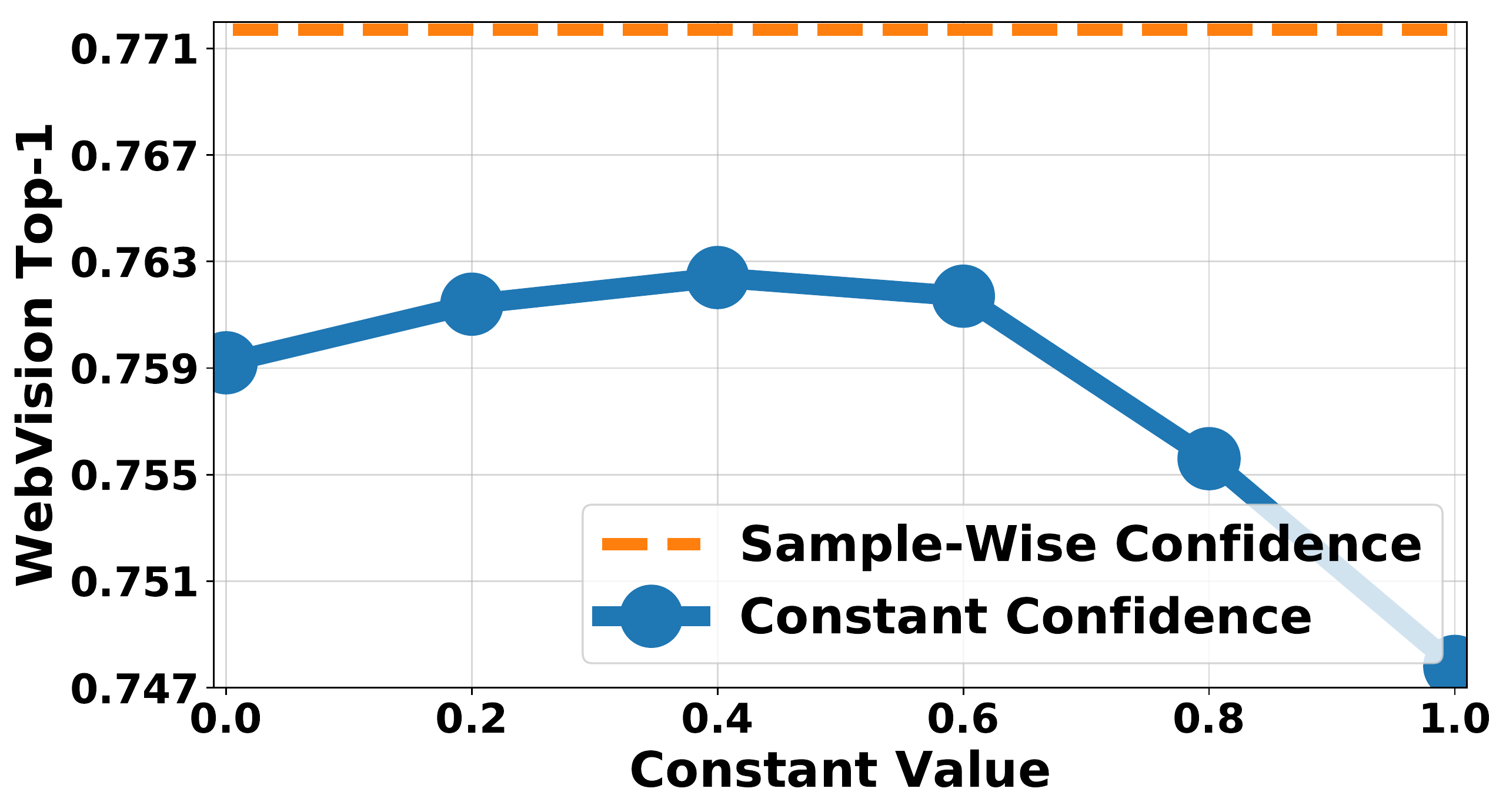}
    \label{fig:ablation-conf}
}
\hfill
\subfloat[Self-label supervised loss]{
    \includegraphics[width=.43\textwidth]{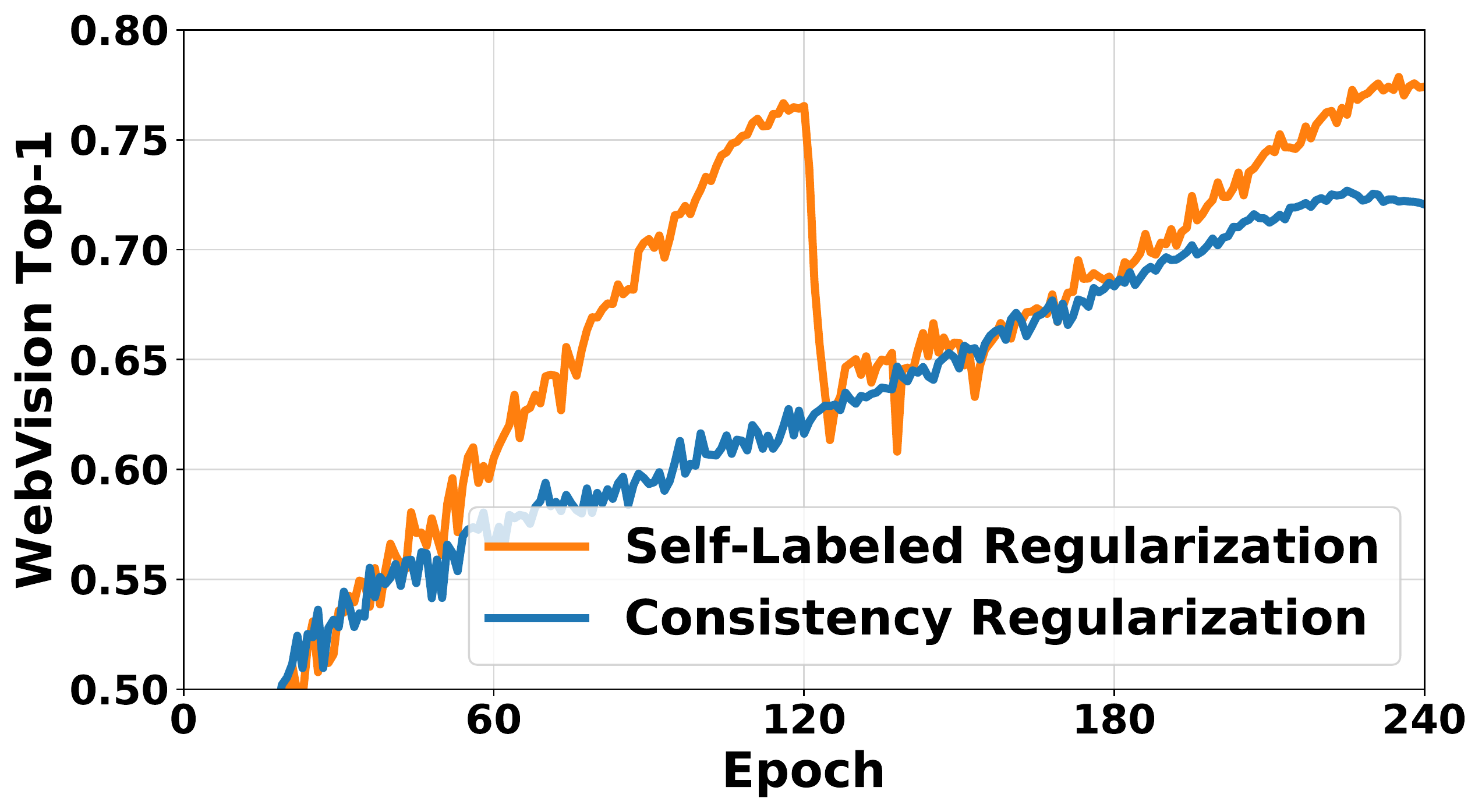}
    \label{fig:ablation-loss}
}
	\caption{Ablation studies of sample-wise confidence and self-label supervised loss}
\label{fig:ablation}
\end{figure}

\textbf{On Self-Contained Confidence}.
To show the necessity of sample-wise SCC, we follow the settings of Table~\ref{tab:comparison} and replace SCC with constant confidence values.
Fig.~\ref{fig:ablation} shows that any constant confidence is unable to surpass $77.17\%$ WebVision Top-1 accuracy reached by Vanilla+GBA~(marked as dashed line).

\textbf{On Self-Label Supervised Loss}.
We demonstrate the superiority of self-label supervised loss over consistency loss~\cite{berthelot2019mixmatch,xie2019unsupervised}. 
Consistency loss is trained in an end-to-end fashion since it does not require a pretrained model $\mathcal{M}_{\theta_0}$, 
whereas our self-label supervised loss expects a two-stage approach with static self labels and SCC.
For fairness, we make comparisons using the same backbone with mixup regularization. Fig.~\ref{fig:ablation-loss} shows the model with our loss reaches better performance than consistency loss. An interesting observation is that a performance drop exists at the beginning of S2 in our method. 
Since S1 is trained with web labels, the model may memorize label noise and result in suboptimal performance. Thus, a large LR is required to destruct the noise-affected S1 model, causing a sudden performance drop with S2. Such a two-stage approach is adopted, because we find the end-to-end approach unsuitable for our method: in the early stage, inaccurate pseudo labels and SCC mislead the model, and in the late stage, the model finally obtains reliable SCC, however, small LR cannot correct the accumulated errors.

\begin{table}[b!]
\caption{The state-of-the-art results on WebVision-1000}
\label{tab:web1000}
	\centering
	\begin{tabular}{llcccc}
		\toprule
		\multicolumn{1}{l}{\multirow{2}{*}{Method}}  & \multicolumn{1}{l}{\multirow{2}{*}{Backbone Network}} & \multicolumn{2}{c}{WebVision} & \multicolumn{2}{c}{ImageNet} \\
		& \multicolumn{1}{c}{} & \multicolumn{1}{c}{Top-1} & \multicolumn{1}{c}{Top-5} & \multicolumn{1}{c}{Top-1} & \multicolumn{1}{c}{Top-5} \\
		\midrule
		MentorNet \cite{jiang2018mentornet} &InceptionResNetV2&72.60&88.90&64.20&84.80\\
		CleanNet \cite{lee2018cleannet} &ResNet50&70.31&87.77&63.42&84.59\\
		CurriculumNet \cite{guo2018curriculumnet} &InceptionV2&72.10&89.20&64.80&84.90\\
		Multimodal \cite{shah2019inferring} & InceptionV3&73.15&89.73&-&-\\
		\midrule
		Initial Vanilla Model & ResNet50-D & 75.08 & 89.22 & 67.23 & 84.09\\
		Ours (Vanilla) & ResNet50-D & 75.36 & 89.38 & 67.93 & 84.77\\
		Ours (Vanilla+GBA) & ResNet50-D & 75.69 & 89.42 & 68.35 & 85.24\\
		\midrule
		Initial Mixup Model & ResNet50-D & 75.54 & 90.36 & 68.77 & 86.59 \\
		Ours (Mixup) & ResNet50-D &75.74&90.78&70.38&88.25\\
		Ours (Mixup+GBA)& ResNet50-D &\textbf{75.78}&\textbf{91.07}&\textbf{70.66}&\textbf{88.46}\\
		\bottomrule
		\noalign{\bigskip}
	\end{tabular}
\end{table}

\subsection{Real-world Experiments}

\textbf{WebVision-1000}.
Table~\ref{tab:web1000} reports experiments on WebVision-1000 using both vanilla and mixup models.
With the vanilla model, using our pipeline, a 0.3\% improvement is achieved for WebVision top-1 accuracy, and 0.7\% increase on ImageNet top-1/5. 
GBA can further improve the performance of every metric.
When enabling mixup operation~($\alpha=0.2$), although on top of a high-accuracy pretrained model, our method can still improve both ImageNet top-1 and top-5 accuracy by 1.9\%. The WebVision top-5 accuracy is improved by 0.7\%. The WebVision top-1 accuracy is improved a little. More improvements on ImageNet prove a good generalization ability of the proposed method. The larger improvement than vanilla may attribute to the higher SCC quality achieved by mixup. GBA advances an average $0.3\%$ extra improvement on every metric.

We also show the superiority of our method over state-of-the-art methods. Note that both MentorNet~\cite{jiang2018mentornet} and CleanNet~ \cite{lee2018cleannet} use extra human-verified datasets to train a guidance network first, and MentorNet~\cite{jiang2018mentornet} chooses a backbone of InceptionResNetV2 stronger than ResNet50-D. Multimodal image classification~\cite{shah2019inferring} uses ImageNet data for training visual embedding and a query-image pairs dataset for training phrase generation. Stronger InceptionV3 is also selected as the backbone. Although with these disadvantages, our ResNet50-D still works the best among all.

\textbf{Food-101N}.
According to Table~\ref{tab:food101N}, we significantly advance the state-of-the-art model without any usage of human annotations.
For vanilla model, the second stage of our method pushes $0.8\%$ higher accuracy than the first stage, and the usage of GBA even double the improvement.
For the mixup model ($\alpha=0.5$), the second stage increases a higher $1.4\%$ accuracy as mixup provides better SCC and self labels than vanilla, but the advance of GBA is deducted due to the overlapping effects of mixup and GBA.
Rather than our normally used ResNet50-D, we use standard ResNet50 here for fair comparisons with others. While all the other methods (except~\cite{zhang2019metacleaner}) train Food-101N from ImageNet pretrained model, we train our model from scratch and still reach optimal performance.

\begin{table}[t!]
\caption{The state-of-the-art results on Food-101N}
\label{tab:food101N}
	\centering
	\begin{tabular}{@{\hskip 5pt}l@{\hskip 30pt}c@{\hskip 5pt}}
		\toprule
		Method & Top-1 \\
		\midrule
		CleanNet \cite{lee2018cleannet} & 83.95\\
		Guidance Learning \cite{li2019product}& 84.20\\
		MetaCleaner \cite{zhang2019metacleaner}& 85.05\\
		Deep Self-Learning \cite{han2019deep} & 85.11\\
		SOMNet \cite{tu2020learning} & 87.50\\
		\midrule
		Initial Vanilla Model &84.08\\
		Ours (Vanilla) & 84.87\\
		Ours (Vanilla+GBA) & 85.76\\
		\midrule
		Initial Mixup Model &86.00\\
		Ours (Mixup) & 87.43\\
		Ours (Mixup+GBA) & \textbf{87.55}\\
		\bottomrule
		\noalign{\bigskip}
	\end{tabular}
\end{table}

\section{Conclusion}
We propose a generic noise-robust framework featured by sample-level confidence balancing webly supervised loss and self-label supervised loss. Our framework is compatible with model regularization methods, among which our proposed mixup+GBA is the most effective. 

Here we recall two main takeaway messages from our extensive experiments:
(1) Reliability of the web label can be reflected by SCC~(ref. Fig.\ref{fig:ece}), and empirical metric SAV is generally proportional to the statistical metrics like ECE~(ref. Table~\ref{tab:comparison}).
(2) Our framework is in favor of high-quality confidence provided by the pretrained model, and mixup and ensemble are the two most effective regularizers~(ref. Fig.\ref{tab:stage1}\&\ref{tab:comparison}). Considering that mixup smooths discriminative spaces and ensemble averages models’ biases, both advantages are combined in GBA design, as a graph smoothing operator for neighbor predictions.

We also leave a valuable discussion in the appendix for readers of interests, which basically shows: although the performance is largely dependent on the quality of SCC, the framework still works on Food101N even with a weak DNN backbone.

\bigskip
\noindent\textbf{Acknowledgement}. 
The work described in this paper was partially supported by Innovation and Technology Commission of the Hong Kong Special Administrative Region, China (Enterprise Support Scheme under the Innovation and Technology Fund B/E030/18).
	
\clearpage
%
%
\bibliographystyle{splncs04}
\bibliography{egbib}

\end{document}